\documentclass[10pt,twocolumn,letterpaper]{article}

\usepackage{wacv}
\usepackage{times}
\usepackage{epsfig}
\usepackage{graphicx}
\usepackage{amsmath}
\usepackage{amssymb}
\usepackage{bbm}
\usepackage{booktabs}
\usepackage{xcolor}
\usepackage{enumerate}
\usepackage{enumitem}
\usepackage{subcaption}

\definecolor{Gray}{gray}{0.5}

\newcommand{\gray}[1]{\textcolor{lightgray}{#1}}

\newcommand{\ytbvis}{YouTube-VIS 2019\xspace}

\newcommand{\trackingbydetection}{\textit{tracking-by-detection}\xspace}
\newcommand{\methodname}{ROVIS}

\usepackage{tabulary,multirow,overpic,xcolor,subfloat}

\def\wacvPaperID{1595} 

\wacvalgorithmstrack   

\wacvfinalcopy 

\ifwacvfinal
\usepackage[breaklinks=true,bookmarks=false]{hyperref}
\else
\usepackage[pagebackref=true,breaklinks=true,colorlinks,bookmarks=false]{hyperref}
\fi

\begin{document}

\title{Robust Online Video Instance Segmentation with Track Queries}

\newcommand{\authorskip}{\hspace{2.5mm}}
\author{Zitong Zhan \authorskip Daniel McKee \authorskip Svetlana Lazebnik \\
University of Illinois at Urbana-Champaign (UIUC)\\
{\tt\small zitongz3@illinois.edu 
\authorskip
dbmckee2@illinois.edu
\authorskip
slazebni@illinois.edu
}
}

\maketitle
\thispagestyle{empty}

\begin{abstract}
Recently, transformer-based methods have achieved impressive results on Video Instance Segmentation (VIS). However, most of these top-performing methods run in an {\em offline} manner by processing the entire video clip at once to predict instance mask volumes. This makes them incapable of handling the long videos that appear in challenging new video instance segmentation datasets like UVO and OVIS. We propose a fully {\em online} transformer-based video instance segmentation model that performs comparably to top offline methods on the YouTube-VIS 2019 benchmark and considerably outperforms them on UVO and OVIS. This method, called {\em Robust Online Video Segmentation (ROVIS)}, augments the Mask2Former image instance segmentation model with {\em track queries}, a lightweight mechanism for carrying track information from frame to frame, originally introduced by the TrackFormer method for multi-object tracking. We show that, when combined with a strong enough image segmentation architecture, track queries can exhibit impressive accuracy while not being constrained to short videos \footnote{Project page:  \url{https://zitongzhan.github.io/rovis_page/}}. 
\end{abstract}


\section{Introduction}

\begin{figure*}
\begin{center}
\includegraphics[width=.95\linewidth]{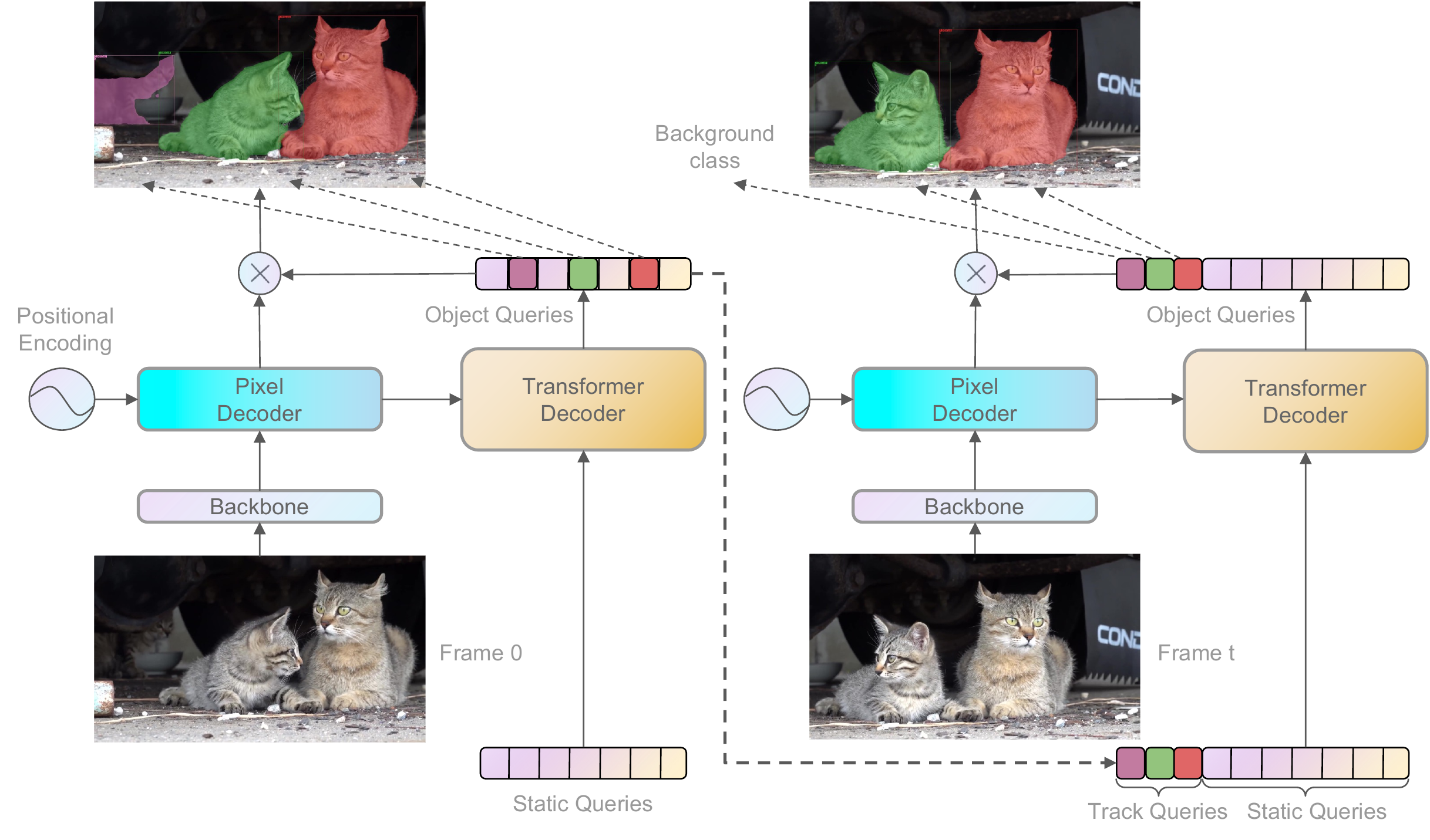}
\end{center}
   \caption{\textbf{System overview.} 
   Our approach is built on the Mask2Former detector which consists of a backbone, pixel decoder, transformer decoder, and class/mask prediction heads. At an initial frame (labeled frame 0), the detector makes object mask predictions based on a set of static \textit{static queries}. Output embeddings from the transformer decoder corresponding to each input query are referred to as \textit{object queries}. The object queries which are decoded to detected objects in a frame are carried forward to subsequent frames as \textit{track queries}. At a subsequent frame $t$ where tracks have previously been established, predictions are made based on the combined set of static and track queries. Track query representations are updated in each frame. When a tracked object disappears, the model predicts a ``background class" label for the track query corresponding to that object as shown in the case of the purple track.
   }
\label{fig:overview}
\end{figure*}

Video instance segmentation (VIS) is a challenging problem that requires detecting, segmenting, and tracking objects through a video. It
emerged as a major benchmark task with the release of the \ytbvis dataset \cite{yang2019video}. 
Initial methods for this task used a \trackingbydetection paradigm where objects are detected at the frame level and then linked across frames to form tracks \cite{fang2021instances,qi2022occluded,yang2019video,crossvis}. 
Such methods mainly operated in an \textit{online} manner, with predictions made immediately for each frame as it appears. But very quickly, a class of {\em offline} transformer-based methods \cite{cheng2021mask2formervis,ifc,vistr,wu2021seqformer} became dominant. These methods take as input the full set of video frames in a clip and predict a mask volume for the video using some form of spatio-temporal attention, feature aggregation, or global optimization. Since YouTube-VIS clips do not exceed 36 frames, offline global inference has been able to excel on them. However, requiring access to all frames in a video to perform inference is impractical in real-time settings. Further, due to the huge GPU memory demands of end-to-end transformer inference on a full video clip, such methods cannot be applied to videos longer than around 50 frames. 
In particular, they cannot easily handle the latest generation of benchmarks like Occluded Video Instance Segmentation (OVIS) \cite{qi2022occluded} and Unidentified Video Objects (UVO) \cite{wang2021unidentified}, whose videos can be over 175 frames long. Besides being longer than YouTube VIS videos, those from the newer benchmarks are also more challenging. While YouTube VIS videos tend to focus on a few target objects in each frame, OVIS features heavy occlusions, and UVO features crowded scenes and diverse objects not restricted to a closed vocabulary, making it more difficult for trackers to correctly link instances across frames. 

To create an online method capable of meeting the demands of the latest benchmarks, we propose a simple yet accurate and efficient combination of up-to-date components from still image segmentation and multi-object tracking (MOT) literature. Our approach,
called ROVIS (for \textbf{R}obust \textbf{O}nline \textbf{V}ideo \textbf{I}nstance \textbf{S}egmentation), starts with the powerful still image segmentation architecture based on Mask2Former~\cite{cheng2021mask2former} and augments it with {\em track queries}, a lightweight mechanism for carrying track information from frame to frame, originally introduced by the TrackFormer MOT method~\cite{meinhardt2021trackformer}.

Figure~\ref{fig:overview} shows an overview of our system. 
In the initial frame, the Mask2Former detector makes object mask predictions based on a set of \textit{static queries} that are fixed regardless of input. The transformer decoder maps these input queries to {\em object queries}, a subset of which correspond to detected objects in the frame and are carried forward to subsequent frames as \textit{track queries}. At a subsequent frame, predictions are made based on the combined set of static queries (for detecting new objects) and existing track queries (to continue tracks of objects from previous frames). Track query representations are updated in each frame; when a tracked object disappears, the model predicts a ``background class" label for the track query corresponding to that object. 


The advantages of our method can be summarized as follows:
\begin{itemize}
\item ROVIS demonstrates that the track query idea can be seamlessly integrated with Mask2Former to effectively perform online VIS. There is no need for a separate linking embedding or optimization step like in other recent online methods, e.g.,~\cite{IDOL}.
    \item ROVIS is efficient for both training and inference. It converges in only 3 hours on four NVIDIA A40 GPUs on the UVO dataset with a ResNet50 backbone, and runs at 12 fps at test time. 
    \item ROVIS reaches accuracies comparable to top offline methods on the YouTube-VIS 2019 benchmark and vastly outperforms those on UVO and OVIS. It also matches or exceeds the performance of other concurrently introduced online methods, IDOL~\cite{IDOL} and MinVIS~\cite{huang2022minvis}.
\end{itemize}





\section{Related Work} \label{sec:related-work}

\noindent\textbf{Video Instance Segmentation.} Initial VIS approaches adapted popular still image models like Mask R-CNN \cite{he2017mask} to video using an online tracking-by-detection approach, in which detections are sequentially predicted for each frame and linked to tracks from previous frames using spatial and appearance similarity. MaskTrack R-CNN \cite{yang2019video} and follow-up works \cite{cao2020sipmask,fang2021instances,qi2022occluded,wang2020solov2,crossvis} are all based on this strategy.

Subsequently, offline methods for VIS aimed at short YouTube-VIS videos became popular and reached top performance on this benchmark. While past offline methods used mask propagation mechanisms \cite{bertasius2020classifying} or graph neural networks for linking \cite{johnander2020learning}, the most recent methods such as VisTR \cite{vistr}, SeqFormer \cite{wu2021seqformer}, IFC \cite{ifc} and Mask2Former VIS \cite{cheng2021mask2formervis} have adopted transformer-based detectors~\cite{carion2020detr,cheng2021mask2former,cheng2021per} to predict tracks in an end-to-end fashion.
They work by adding a temporal input dimension to predict 3D mask volumes for full videos or subclips. This means that tracking is performed by spatio-temporal attention inside the transformer. 
In our approach, track queries take the place of expensive spatio-temporal attention.

Most recently, motivated by similar considerations of efficiency as ourselves, other researchers have started exploring online transformer-based methods for VIS. In our work, we cite and compare performance to two methods, IDOL~\cite{IDOL} and MinVIS~\cite{huang2022minvis}. Both of these are tracking-by-detection methods that start by running a strong per-frame segmentation model and then link instances across frames in different ways. IDOL~\cite{IDOL}, for ``In Defense of Online Models for VIS,'' proposes a contrastive objective for learning of discriminative instance embeddings for association, and at test time, it uses a memory-bank-based linking scheme with the learned embeddings on top of per-frame segmentations. In our experiments, our proposed method reaches similar levels of performance as IDOL on YouTube-VIS and OVIS, and outperforms it on UVO, while having a faster inference time due to seamlessly carrying track information from frame to frame using track queries, without requiring a separate feature representation for linking and a linking optimization step. The second online method, MinVIS~\cite{huang2022minvis}, for ``Minimal VIS,'' also starts by performing per-frame instance segmentation. In fact, it is based on the same Mask2Former per-frame model that we use. Then, during the tracking stage, for each pair of neighboring frames, MinVIS performs bipartite matching on the sets of object queries produced in per-frame detection of Mask2Former. The advantage of this scheme is that it is simple and does not require any video-specific training. As will be shown in Section \ref{sec:main-results}, it achieves impressive results on the YouTube-VIS dataset, but underperforms both IDOL and our method on OVIS. 



\vspace{0.5em}
\noindent\textbf{Multi-Object Tracking Methods.} The VIS task is closely related to the more traditional multi-object tracking (MOT) task where objects are detected and tracked but not segmented (though recently, segmentation has also been added to some MOT datasets). For historical reasons, there are application domain differences between VIS and MOT: MOT datasets generally focus on autonomous driving \cite{geiger2013vision,yu2020bdd100k} and person tracking \cite{MOT16,dendorfer2020mot20}, while VIS datasets focus on more semantically diverse videos \cite{yang2019video,qi2022occluded,du2021uvo} aligned with popular multi-category datasets like COCO \cite{lin2014microsoft}.
While there exists work attempting to unify VIS and MOT \cite{athar2020stem}, current top methods are task-specific to either VIS or MOT.


Past approaches to MOT have primarily followed the tracking-by-detection paradigm, encompassing both offline methods \cite{andriyenko2011multi, berclaz2006robust,braso2020learning,jiang2007linear, pirsiavash2011globally, tang2017multiple, xiang2015learning, xu2019spatial,zhang2008global}, which typically involve a graph optimization process, and online methods \cite{bewley2016simple,pang2021quasi,wang2019towards,wojke2017simple,zhang2021bytetrack,zhang2021fairmot}, which associate detections to previous tracks on the fly using various spatial or appearance similarity measures. More recent ``regression-based'' models \cite{bergmann2019tracking,feichtenhofer2017detect,mckee2021multi,shuai2021siammot,zhou2020tracking} predict positions of objects from previous frames to perform data association. 

The MOT approaches most relevant to our work are a new class of methods jointly performing detection and tracking using an end-to-end transformer with track queries \cite{meinhardt2021trackformer,sun2020transtrack,zeng2021motr}. We investigate a track query scheme for VIS most similar to that of TrackFormer~\cite{meinhardt2021trackformer}. The original TrackFormer 
model, based on a vanilla DETR~\cite{carion2020detr} transformer design, mainly targeted MOT challenges but also provided a segmentation model applicable to VIS. It should be noted that TrackFormer is no longer among the strongest methods on MOT since accuracy on MOT at present seems to mainly depend on detection quality. The leaderboard for MOT 20~\cite{dendorfer2020mot20} is dominated by methods such as \cite{zhang2021bytetrack} that use the strong CNN-based YOLO-X detector~\cite{yolox2021}, which considerably surpasses DETR on COCO object detection; as shown in~\cite{zhang2021bytetrack}, matching bounding boxes based on IoU is already sufficient for associating detected objects in MOT benchmarks. In Section \ref{sec:main-results}, we will compare our model against the original TrackFormer on VIS and show that combining the track query idea with the Mask2Former backbone yields dramatically better results.

\vspace{0.5em}
\noindent\textbf{Open-World and Occluded Tracking Datasets.} With the maturation of tracking and VIS tasks, recently introduced datasets have started to emphasize more challenging scenarios like tracking of objects of unknown category and tracking through heavy occlusions. 
The TAO-OW dataset \cite{liu2022opening} builds an open-world MOT benchmark on top of the long-tailed TAO dataset \cite{dave2020tao} by introducing a set of ``unknown" class labels which are not included in the training set. Both TAO and TAO-OW are not exhaustively labeled.
In contrast, the UVO dataset \cite{du2021uvo}, on which we focus our open-world study, offers VIS task annotations by providing segmentation masks for all object instances and handles open-world tracking by exhaustively labeling all object instances in a class-agnostic manner for a subset of 1.2k Kinetics-400 \cite{kay2017kinetics} videos.
In addition to open-world tracking, challenging closed-world settings with heavy occlusions have also garnered interest with the release of the OVIS dataset \cite{qi2022occluded} in 2021.

We are most interested in tackling these challenging open-world and occluded scenarios during tracking. For the UVO and OVIS datasets, we found that existing methods are inferior on either single-frame segmentation performance or are unable to handle complex tracking situations. For example, the OVIS dataset contains large texture deformations between frames and objects disappearing and reappearing. Offline methods fail to handle such cases even though they work well on common VIS benchmarks, as we will show in our experiments.

\begin{figure*}[t]
  \centering
   \includegraphics[width=.95\linewidth]{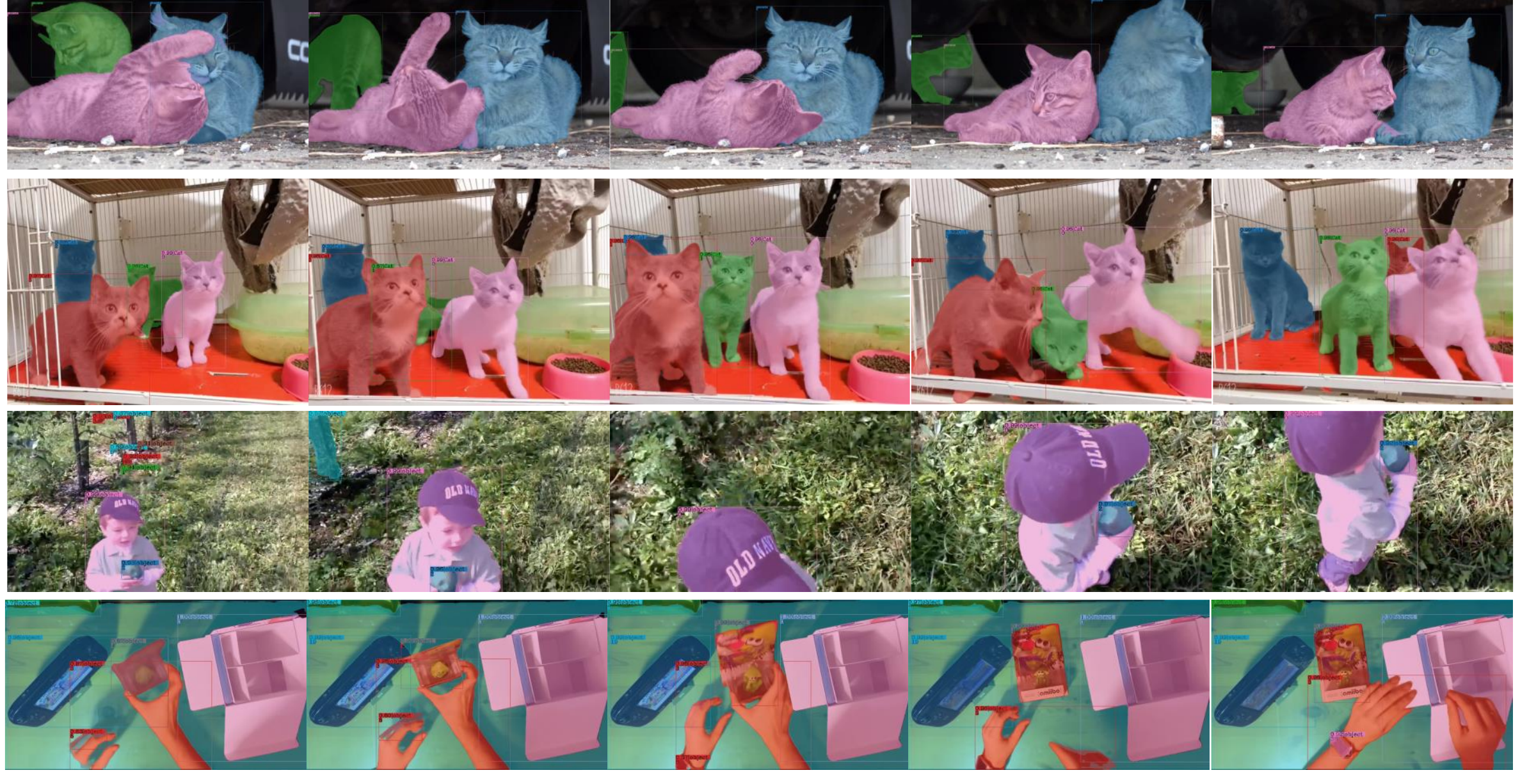}

   \caption{Qualitative results for our Swin-L model on OVIS validation (top two rows) and UVO \texttt{dense-val} (bottom two rows). Selected frames are uniformly spaced. Our model performs well under cases of significant crowding and occlusions in OVIS. 
   In the second row, the foreground ``red'' kitten in frames 1-4 rapidly moves behind the ``green" and ``pink" ones to end up in the background in the last frame.
   On UVO, our model performs well with significant offscreen occlusion (apple held by the child in the third row) and appearance deformation (packet in the bottom row).}
   \label{fig:quals_ovis_uvo}
\end{figure*}



\label{sec:method}
\section{Method}
This section presents the details of our ROVIS method.
Section~\ref{subsection:method_arch} will discuss our per-frame instance segmentation architecture and track queries. Section~\ref{subsection:loss} will discuss the training scheme and losses, and Section~\ref{subsection:inference} will discuss inference.

\subsection{Method Architecture} \label{subsection:method_arch}

Figure \ref{fig:overview} shows an overview of our method.
We choose to build on the powerful Masked-attention Mask Transformer (Mask2Former) still image segmentation architecture of Cheng et al.~\cite{cheng2021mask2former}. 
The major components of Mask2Former are a backbone, a pixel decoder, and a transformer decoder. The backbone extracts a feature map from the input frame, which the pixel decoder gradually modifies and upsamples across multiple levels. At each level, the transformer decoder takes in a feature map from the pixel decoder along with a trained set of queries, and performs {\em masked cross-attention} between them, i.e., cross-attention restricted to a predicted subset of feature locations. The output of each level of the transformer decoder is a set of modified query embeddings that are fed into the next decoder level. The object queries and the feature map output by the final level of the decoders are combined to make final mask and class predictions for object instances. Masks are produced by pixel-wise dot product between image features from a convolutional mask head and object queries, and class predictions are produced by linear layers connected directly to the object queries. Aside from the object categories in the target dataset, a ``background'' class is added to the possible outputs of the class prediction head to represent the case when no instance is found by the corresponding query. In the class-agnostic UVO dataset, there is a single ``object'' class assigned to all instances, so the detector prediction is either ``object'' or ``background'' when no object is found. 

In our tracking scenario, we equate the pre-trained queries that are fed into the first level of the decoder with {\em static queries} that are used to initialize tracks in each frame. The ``successful'' object queries that emerge from the last level of the decoder (i.e., the queries that predict object instances) are used to carry instance information from frame to frame. Accordingly, as shown in Figure \ref{fig:overview} (right), after the first frame, we augment the static queries with a variable-size set of {\em track queries} initialized from the object queries associated with predicted instances in the previous frame. 
As we will show, both types of queries can be trained to retrieve object appearance information from the pixel decoder effectively, allowing consistent instance segmentation across frames.

\subsection{Training Procedure and Losses} \label{subsection:loss}

\noindent \textbf{Instance mask prediction and losses.} As described above, given a single frame, for each input query, the Mask2Former detector outputs a class and binary mask prediction. 
Class prediction is supervised by the cross-entropy loss $\mathcal{L}_{\text {cls}}$ and masks are supervised by the mask loss $\mathcal{L}_{\text {mask}}$. Following Mask2Former \cite{cheng2021mask2former}, for each predicted mask and ground truth, we calculate \textit{focal loss}~\cite{focalloss} and \textit{dice loss}~\cite{dicelossMilletari2016VNetFC} at $K$ sampled locations, and the full mask prediction loss is $\mathcal{L}_{\text {mask}} = \mathcal{L}_{\text {focal}} + \mathcal{L}_{\text {dice}}$. The classification loss is applied to all predictions, while the mask loss is only applied to predictions that correspond to a non-background instance. 



\vspace{0.5em}
\noindent\textbf{Training for end-to-end tracking}. Our training framework is fully supervised, i.e., it relies on ground-truth instance IDs. 
With this supervision, track queries learn to detect and fire on the same instances that they represented in previous frames.
Following \cite{cheng2021mask2formervis} and \cite{meinhardt2021trackformer}, each iteration of the training process involves only a pair of nearby frames randomly sampled from a video. We denote these frames $x^0$ and $x^1$, respectively. 
Instead of always choosing $x^1$ as the frame immediately after $x^0$, we uniformly sample $x^1$ within a range of frames neighboring frames around $x^0$ during training. We found that a range of $\pm$15 frames works best for UVO and $\pm$5 works best for the rest of the datasets. Note that $x^1$ can actually be a frame earlier than $x^0$, which allows the network to be trained on reversed motions as well. 

Given a pair of frames, training proceeds as follows:
\begin{enumerate}
   \item Perform prediction in $x^0$ using static queries (this corresponds to considering all instances in $x^0$ to be new objects). Use the Hungarian algorithm to compute an assignment of predictions to ground truth instances. The assignment cost function for matching a prediction to the ground truth is defined exactly the same as the detector prediction loss, $\lambda_{\text {cls }} \cdot\mathcal{L}_{\text {cls}} + \lambda_{\text {mask }} \cdot\mathcal{L}_{\text {mask}}$ (see eq.(\ref{eq:loss}) for details of the constants). Predictions not assigned to any ground-truth instance are assigned the ``background'' label. 
   Given the matching, compute the classification and mask prediction losses for $x^0$, denoted as $\mathcal{L}_{\text {cls}}^0$ and $\mathcal{L}_{\text {mask}}^0$. 
   \item Extract all object queries associated with the ground-truth instances in $x^0$ and concatenate them to form track queries for $x^1$. Assign the track queries in $x^1$ to predict the same instances as the corresponding object queries in $x^0$. For ground-truth instances in $x^0$ that disappear in $x^1$, assign the track query to predict the background class. Compute the losses for tracked instances in $x^1$, denoted as $\mathcal{L}_{\text {cls}}^{\text{track}}$ and $\mathcal{L}_{\text {mask}}^{\text{track}}$. 
   \item For any new instances that appear in $x^1$ (i.e., any instances other than the ones tracked from $x^0$), use greedy assignment to match their ground-truth masks to mask predictions produced by static queries. Compute losses for new objects in $x^1$, denoted as $\mathcal{L}_{\text {cls}}^1$ and $\mathcal{L}_{\text {mask}}^1$.
\end{enumerate}
Finally, the combined loss for each iteration is
\begin{equation} \label{eq:loss}
\mathcal{L}
=
\lambda_{\text {cls }} ( \mathcal{L}_{\text {cls}}^0 + \mathcal{L}_{\text {cls}}^{\text{track}} + \mathcal{L}_{\text {cls}}^1) 
+ \lambda_{\text {mask }} (\mathcal{L}_{\text {mask}}^0 + \mathcal{L}_{\text {mask}}^{\text{track}} + \mathcal{L}_{\text {mask}}^1)
\end{equation}
\noindent where $\lambda_{\text {cls}}$ and $\lambda_{\text {mask}}$ are set to 2.0 and 5.0 in all experiments.

\vspace{0.5em}
\noindent\textbf{Training-time track augmentation.} We rely on a query augmentation strategy modified from TrackFormer~\cite{meinhardt2021trackformer}.
At inference time, if the model fails to detect a certain object in frame $t-1$, the static queries should remain responsible for detecting it in frame $t$. To simulate this circumstance during training, track queries have a probability of $p_{FN} = 0.2$ of being removed from training frame $x^1$.
The corresponding ground truth object for a dropped track query in frame $x^1$ will instead be matched with static queries, designed to trigger a new object detection. 
To simulate termination and occlusion of tracks, each object query from $x^0$ that predicts the background class has probability $p_{FP}=0.2$ of being added to the set of track queries. These false positive queries will be trained to predict background class in $x^1$. Removing the false positive augmentation 
causes roughly a 3-point AP drop on the OVIS dataset, and removing the false negative augmentation causes a 2-point drop.

\begin{table*} 
\centering
\begin{tabular}{llllccccc}
\toprule
& Method & Backbone & Type & AP & AP$_{50}$ & AP$_{75}$ & AR$_{1}$ & AR$_{10}$\\
\midrule
\multicolumn{9}{l}{(a) Previous methods}\\
\midrule
& MaskTrack R-CNN~\cite{yang2019video} & ResNet50  &{online}  &30.3 &51.1 &32.6 &31.0 &35.5  \\ 
& CrossVIS~\cite{crossvis}     & ResNet50  &{online}  &36.3 &56.8 &38.9 &35.6 &40.7   \\
 
& IFC~\cite{ifc}   & ResNet50  &\gray{offline}   &42.8 &65.8 &46.8 &43.8 &51.2  \\  
& {SeqFormer}~\cite{wu2021seqformer} & ResNet50  &\gray{offline}    &45.1 &66.9 &50.5 & {45.6} &54.6  \\
& Mask2Former VIS \cite{cheng2021mask2formervis} & ResNet50 &\gray{offline}& {46.4} & 68.0 & 50.0 & - & - \\
\midrule
\multicolumn{9}{l}{(b) Baselines and our method}\\
\midrule
& Mask2Former + IoU & ResNet50 & {online}& {37.9} & 55.3 & 41.3 & 37.0 & 41.9 \\
& {TrackFormer} & ResNet50  & {online}    &19.8 &34.4 &20.9 & {20.3} & 25.2  \\
& \methodname\xspace (Ours) & ResNet50 &{online}&  45.5 & 63.9 & 50.2 & 41.8 & 49.5\\
\midrule
\multicolumn{9}{l}{(c) Concurrent state-of-the-art methods}\\
\midrule
& IDOL~\cite{IDOL}  & ResNet50  &{online}  & {46.4} & \textbf{70.7} & {51.9} & {44.8} &  {54.9} \\  
& MinVIS~\cite{huang2022minvis}  & ResNet50  &{online}  & \textbf{47.4} & {69.0} &\textbf{52.1} & \textbf{45.7} &\textbf{55.7} \\  
\bottomrule
\end{tabular}
\caption{Results on \ytbvis validation set when no additional training data is used.}
\label{ytvis-main}
\end{table*}

\subsection{Inference} \label{subsection:inference}
\noindent\textbf{Track inactivity tolerance.} Similar to MaskTrack R-CNN \cite{yang2019video}, after an instance disappears in a new frame of a video, we wait until it has been missing for more than $\Delta t$ frames before removing the corresponding track query ($\Delta t = 9$ in our experiments). This allows the system to preserve the original track id after an object temporarily disappears rather than assigning the reappeared instance an incorrect new id. 

\vspace{0.5em}

\noindent\textbf{Non-maximum suppression (NMS).} 
Although the transformer decoder is trained to avoid producing duplicate instances, in practice different track queries can end up predicting the same instance. Thus, we apply NMS to only keep the track that produces the mask with a higher confidence. 
We use matrix NMS \cite{wang2020solov2} in our YouTube VIS and OVIS experiments since these datasets have category annotation, and plain NMS in our UVO experiments since the UVO annotations are class-agnostic.
NMS is first performed on the instances proposed by track queries only, and then on both types of instances jointly.
Removing NMS causes a 2$\%$ accuracy drop on OVIS \texttt{val}.

\section{Experiments}
In Section \ref{sec:implementation} we discuss the details of our implementation, and
in Section \ref{sec:main-results} we compare our model with the state of the art on YouTube-VIS 2019 \cite{yang2019video}, UVO \cite{wang2021unidentified}, and OVIS \cite{qi2022occluded} datasets. 

\subsection{Implementation Details} \label{sec:implementation}
We build our pipeline based on MMDetection \cite{mmdetection} and MMTrack \cite{mmtrack2020} toolboxes. Training is performed using the AdamW~\cite{loshchilov2018decoupled} optimizer with learning rate and weight decay set to $2.5 \times 10^{-5}$ and $0.05$, respectively. The detector is initialized with weights pre-trained on COCO instance segmentation. The backbone has a learning rate multiplier of 0.1, and weight decay is not performed for the static queries or class embeddings. 
We train our model for 3 epochs on the UVO and YouTube-VIS dataset regardless of backbone architecture. For OVIS, the ResNet50 \cite{he2016deep} backbone model is trained for 4 epochs, and the Swin-L \cite{liu2021swin} backbone model is trained for 3 epochs. During training and inference, frames are resized so that the shorter side length is 480 pixels. None of our experiments use additional training data. The ResNet50 variant of our model has 100 static queries, and the Swin-L variant has 200 static queries, for all datasets.

We report all results using the standard metrics of average precision (AP) and average recall (AR). AP is computed at 10 intersection-over-union (IoU) thresholds, from
50\% to 95\% with 5\% step size, then averaged to obtain the mean value. IoU calculation is based on 3D volumetric IoU \cite{yang2019video}.

\subsection{Main Results} \label{sec:main-results}

\noindent{\bf YouTube-VIS results.}
Table \ref{ytvis-main} shows a comparison of ResNet50 backbone methods on YouTube-VIS 2019 \cite{yang2019video}, which is the most mature and established benchmark in our experiments, consisting of relatively short videos with few objects and simple motions. For completeness, Table \ref{ytvis-main}(a) lists earlier online methods like MaskTrack R-CNN \cite{yang2019video} and CrossVIS \cite{crossvis} based on Mask R-CNN \cite{he2017mask} image segmentation, as well as representative offline transformer-based approaches. The latter all get a significant improvement over the online methods, with the best one, at 46.4 AP, being Mask2Former VIS. It is a 3D extension of Mask2Former that performs spatiotemporal attention on all frames at once, and as such, it forms a natural offline reference point for our method.

Table \ref{ytvis-main}(b) reports a couple of our own baselines followed by our ROVIS method. Our first baseline, labeled ``Mask2Former+IoU," is given by the still-image Mask2Former detector with a simple IoU-based heuristic for tracking. Specifically, each instance found in a new frame is linked to the instance in the previous frame whose mask has the highest IoU with it. This baseline achieves 37.9 AP, showing the gap between a 3D transformer and a 2D per-frame one with rudimentary tracking. 
Since our track query idea comes from TrackFormer, we also evaluate the TrackFormer model trained until convergence using settings from the original work~\cite{meinhardt2021trackformer}, and the performance is disappointing at 19.8 AP. On all three VIS datasets we tried, we found the original TrackFormer to have worse performance than methods specifically designed for VIS, no matter whether we used weights pre-trained on COCO panoptic segmentation or MOTS 20~\cite{voigtlaender2019mots}. This confirms that TrackFormer's detector, based on vanilla DETR~\cite{carion2020detr}, is too weak to handle the diverse appearance in VIS, and that the track query idea needs to be applied on top of a strong detector to realize its potential. 
Mask2Former is more optimized than DETR for segmentation performance: although both DETR and Mask2Former stack multiple transformer decoder layers together, DETR only learns segmentation through the output of the last layer, while Mask2Former has segmentation head on each decoder layer being supervised. This limitation of DETR cannot be easily circumvented due to memory cost.



\begin{table*}[t]
\centering
\begin{tabular}{llllccccc}
\toprule
& Method & Backbone & Type & AP & AP$_{50}$ & AP$_{75}$ & AR$_{1}$ & AR$_{10}$ \\
\midrule
\multicolumn{9}{l}{(a) Previous methods}\\
\midrule
& MaskTrack R-CNN \cite{yang2019video} & ResNet50   &{online}  &10.8 &25.3 &8.5 &7.9 &14.9  \\
& CMaskTrack R-CNN~\cite{qi2022occluded} & ResNet50   &{online}  &{15.4} &{33.9} &13.1 &9.3 &20.0  \\
& CrossVIS~\cite{crossvis}   & ResNet50     &{online}  &14.9 &32.7 &12.1 &10.3 &19.8   \\
& {IFC~\cite{ifc}} (as reported in~\cite{IDOL}) & ResNet50  &\gray{offline}  &13.1 &27.8 &11.6 &9.4 & 23.9 \\
& {SeqFormer~\cite{wu2021seqformer}} (as reported in~\cite{IDOL}) & ResNet50  &\gray{offline} &15.1 &31.9 &{13.8} &{10.4} &{27.1} \\
\midrule
\multicolumn{9}{l}{(b) Baselines and our method}\\
\midrule
& {Mask2Former VIS (per-chunk)} & ResNet50  &\gray{offline} & 16.5 & 36.5 & 14.6 & 10.2 & 23.4 \\
& {Mask2Former VIS (CPU)} & ResNet50  &\gray{offline} & 16.6 & 36.9 & 14.1 & 9.9 & 24.7 \\
& {Mask2Former + IoU} & ResNet50  &{online} & 20.9 & 37.2 & 20.8 & 11.1 & 31.2 \\
& {TrackFormer} & ResNet50  &{online} & 8.9 & 18.3 & 8.3 & 4.9 & 15.8 \\
& \methodname\xspace & ResNet50 &{online}&  \textbf{30.2} & \textbf{53.9} & \textbf{30.1} & 13.6 & 36.3\\
\midrule
\multicolumn{9}{l}{(c) Concurrent state-of-the-art methods}\\
\midrule
& {IDOL} (high resolution)~\cite{IDOL} & ResNet50   &{online}   &\textbf{30.2} & 51.3 & 30.0 &\textbf{15.0} &{37.5}  \\  
& {IDOL} & ResNet50 &{online}& 28.2 & 51.0 & 28.0 & 14.5 & \textbf{38.6} \\  
& {MinVIS}~\cite{huang2022minvis} & ResNet50   &{online}   &{25.0} & 45.5 & 24.0 & {13.9} & {29.7}  \\  
\midrule
\multicolumn{9}{l}{(d) SWIN-L backbone comparison}\\
\midrule
& \methodname\xspace (high resolution) & Swin-L &{online} & \textbf{42.6} & 64.7 & 42.6 & {18.4} & 49.1 \\
& \methodname\xspace & Swin-L &{online} & 41.6 & 65.0 & 42.9 & \textbf{18.7} & 46.9 \\
& {IDOL (high resolution)~\cite{IDOL}}  & Swin-L  &{online}   &\textbf{42.6} &\textbf{65.7} &\textbf{45.2} & 17.9 &\textbf{49.6} \\
& {IDOL}  & Swin-L  &{online}  & 40.0 & 63.1 & 40.5 & 17.6 & 46.4 \\
& {MinVIS~\cite{huang2022minvis}}  & Swin-L  &{online}  & 39.4 & 61.5 & 41.3 & 18.1 & 43.3 \\
\bottomrule
\end{tabular}
\caption{
\textbf{Results on OVIS validation set.} ``High resolution'' refers to training and evaluation at 720px; in all other cases, the resolution is 480px. Note that in the published work on IDOL~\cite{IDOL}, only high-resolution results are reported. The low-resolution IDOL results are obtained by us using their released implementation. }
\label{ovis-main}
\end{table*}

As reported in the third line of Table \ref{ytvis-main}(b),
our method achieves an AP of 45.5, just below Mask2Former VIS. Finally, Table \ref{ytvis-main}(c) reports results of two concurrent methods discussed in Section \ref{sec:related-work}, IDOL~\cite{IDOL} and MinVIS~\cite{huang2022minvis}. IDOL is on par with or slightly better than Mask2Former VIS, and MinVIS is slightly higher in turn, at 47.4 AP. Both models adopt a tracking-by-detection design, which is potentially less vulnerable than track queries to propagation of errors from previous frames. Also, in general, YouTube-VIS 2019 is a fairly mature benchmark on which a number of highly tuned methods perform quite similarly. As we will see below, our method outperforms MinVIS on OVIS and IDOL on UVO.


\begin{figure*}[t]
  \centering
   \includegraphics[width=.95\linewidth]{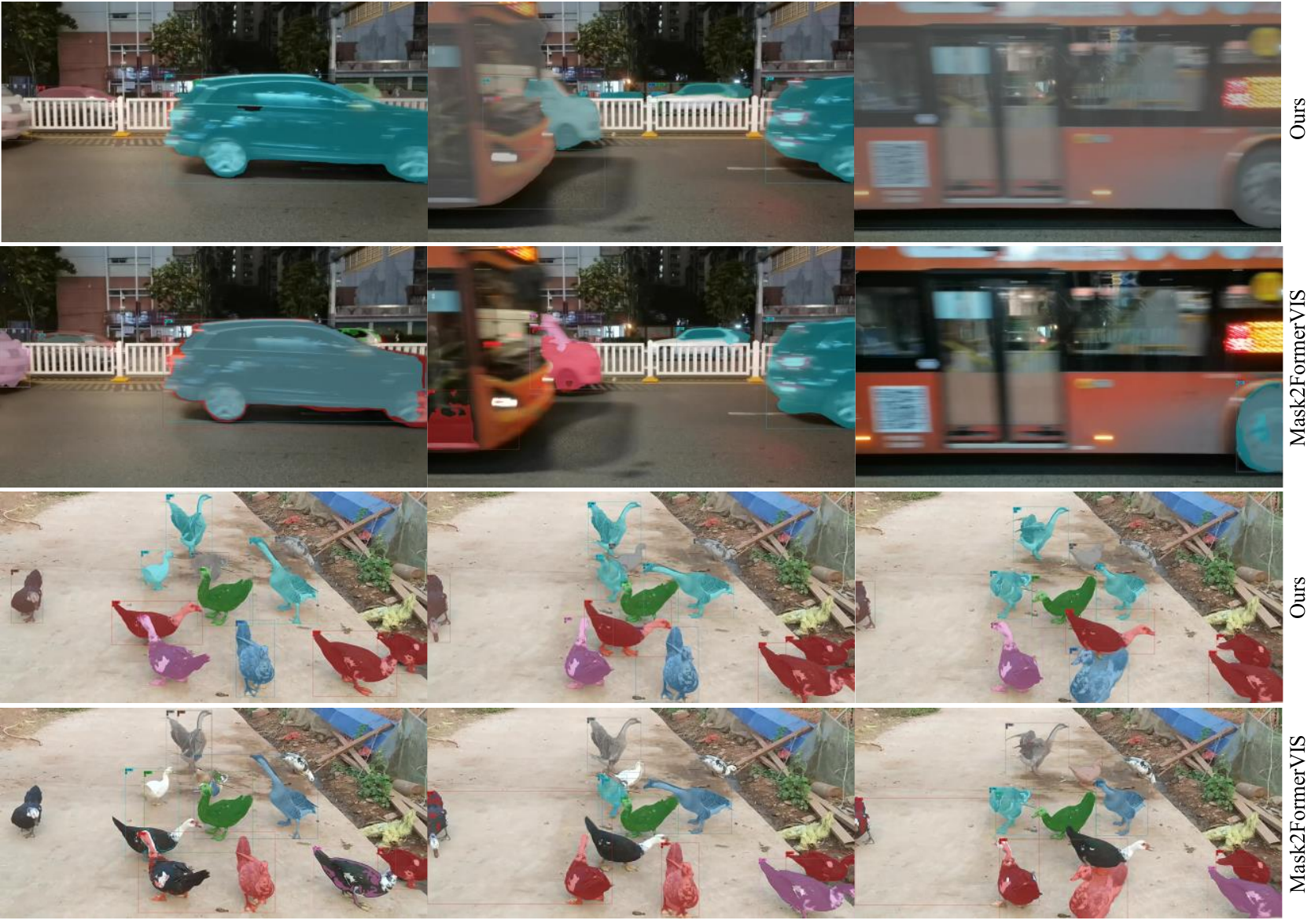}

   \caption{\textbf{Qualitative comparison with Mask2Former VIS on OVIS dataset.} Both models use ResNet-50 backbone. \textbf{Top:} In this fast moving scene with heavy motion blur and occlusion, Mask2Former VIS makes an incorrect association. The wheel of the bus is linked to an SUV from several previous frames. Frames 27, 30, and 36 are selected for visualization. \textbf{Bottom:} In this example with occlusion and crowding, Mask2Former VIS fails to segment numerous instances. Frames are selected with spacing of 4 to better show motion.}
   \label{compare:mask2vis_on_ovis}
\end{figure*}
\noindent {\bf OVIS results.}
Table~\ref{ovis-main} reports results on the OVIS dataset~\cite{qi2022occluded}. 
This dataset consists of 901 total videos (154 in the test set) with an average duration of 12.77 seconds {and up to 500 frames}, annotated at 5-6 fps. OVIS contains frequent object occlusion and deformation, making it difficult for previous methods: as shown in part (a) of the table, the absolute AP numbers are much lower than on YouTube-VIS. Since OVIS includes longer videos, it is not possible for offline transformer methods to process the full video in GPU memory. For IFC \cite{ifc} and SeqFormer \cite{wu2021seqformer}, Wu et al.~\cite{IDOL} report results obtained by processing sub-clips from the videos and linking these predictions. Interestingly, they get no performance improvement over the previous top online methods.

Since there are no previously reported Mask2Former VIS results on OVIS, in Table~\ref{ovis-main} (b), we evaluate it using two different strategies.
 The Mask2Former VIS model is trainable on the GPU as it only takes in two frames at a time, but at inference time it takes in all frames of a video at once, so it cannot handle more than around 50 frames on the GPU. To circumvent the GPU memory limitation, we first evaluate Mask2Former VIS using CPU alone, labeled ``Mask2Former VIS (CPU)." Needless to say, this evaluation is very slow, and the results are barely higher than those of SeqFormer.  
 To make sure that clip length is not the main factor behind this mediocre performance, we also experimented with testing Mask2Former VIS using a per-chunk inference scheme similar to \cite{huang2022minvis}. We split videos in the dataset into smaller chunks of 30 frames, with neighboring chunks overlapping by three frames. Then Mask2Former VIS is used to predict masks volumes for each chunk independently, and per-chunk predictions are linked based on the IoU of the masks within the overlapping frames, analogously to our Mask2Former+IoU baseline. This scheme, labeled ``Mask2Former VIS (per-chunk)" produces accuracy similar to CPU inference, and both are outperformed by 5 AP points by the Mask2Former+IoU baseline. Interestingly, OVIS is the only dataset that causes Mask2Former VIS to perform worse than Mask2Former+IoU, suggesting that the training scheme of Mask2Former VIS is particularly vulnerable to occlusion. ROVIS improves over Mask2Former+IoU by a further 10 AP points, reaching 30.2 AP. Figure~\ref{compare:mask2vis_on_ovis} shows a qualitative comparison between the output of our method and that of Mask2Former-VIS on a couple of sequences.

Table~\ref{ovis-main} (c) gives the reported results for IDOL~\cite{IDOL}, which was specifically developed with OVIS in mind. The 30.2 AP of IDOL exactly matches ours. However, it is important to note the experiments of~\cite{IDOL} use a higher training and evaluation resolution of 720px. For a fairer evaluation, we also trained and evaluated IDOL at 480px resolution, resulting in a drop of around 2 points AP. 
Table~\ref{ovis-main} (c) also lists the reported results of MinVIS, which are appreciably below both ours and IDOL's.
Finally, Table~\ref{ovis-main} (d) shows results for ROVIS, IDOL, and MinVIS with the SWIN-L backbone. Here, the high-resolution versions of ROVIS and IDOL get the best results of 42.6 AP, and MinVIS is once again behind. 

While optimizing ROVIS inference for real-time performance was not our priority, our GPU implementation does run reasonably fast. Our measured inference speed is \textbf{12 fps} for the ResNet-50 backbone and \textbf{5 fps} for the Swin-L backbone on an NVIDIA A40 GPU. By comparison, when running the Swin-L IDOL model on OVIS at 480px resolution, we measured a 3.2 fps inference speed on the same hardware. 
\smallskip

\begin{table*} 
\centering
\begin{tabular}{llllcccc}
\toprule
& Method & Backbone & Type & AP & AP$_{50}$ & AP$_{75}$ & AR$_{100}$ \\
\midrule
\multicolumn{8}{l}{(a) ResNet50 backbone comparison} \\
\midrule
& MaskTrack R-CNN (as reported in \cite{wang2021unidentified}) & ResNet50 & online & 9.3 & 20.9 & 8.2 & 17.2 \\
& Mask2Former VIS (CPU) & ResNet50 & \gray{offline} & 19.0 & \textbf{32.8} & 18.9 & 26.4 \\
& Mask2Former + IoU & ResNet50 & online & 18.2 & 29.7 & 18.9 & 23.6 \\
& IDOL & ResNet50 & online & 16.8 & 28.1 & 17.3 & 23.9 \\
& \methodname\xspace (Ours) & ResNet50 & online  & \textbf{21.1} & \textbf{32.8} & \textbf{21.3} & \textbf{29.1}\\
\midrule
\multicolumn{8}{l}{(b) SWIN-L backbone comparison} \\
\midrule
& Mask2Former VIS (CPU) & Swin-L & \gray{offline}  & 27.3 & 42.0 & 27.2 & 35.4\\
& \methodname\xspace (Ours) & Swin-L & online  & \textbf{32.7} & \textbf{46.4} & \textbf{33.3} & \textbf{41.2}\\
\bottomrule
\end{tabular}
\caption{\textbf{Results on UVO \texttt{dense-val} split.} All models are trained on the UVO \texttt{dense-train} split. Since Mask2Former VIS cannot process videos over 50 frames on GPU due to memory constraints, we run inference using CPU compute only. }
\label{uvo-main}
\end{table*}

\begin{figure*}[t]
  \centering
   \includegraphics[width=.95\linewidth]{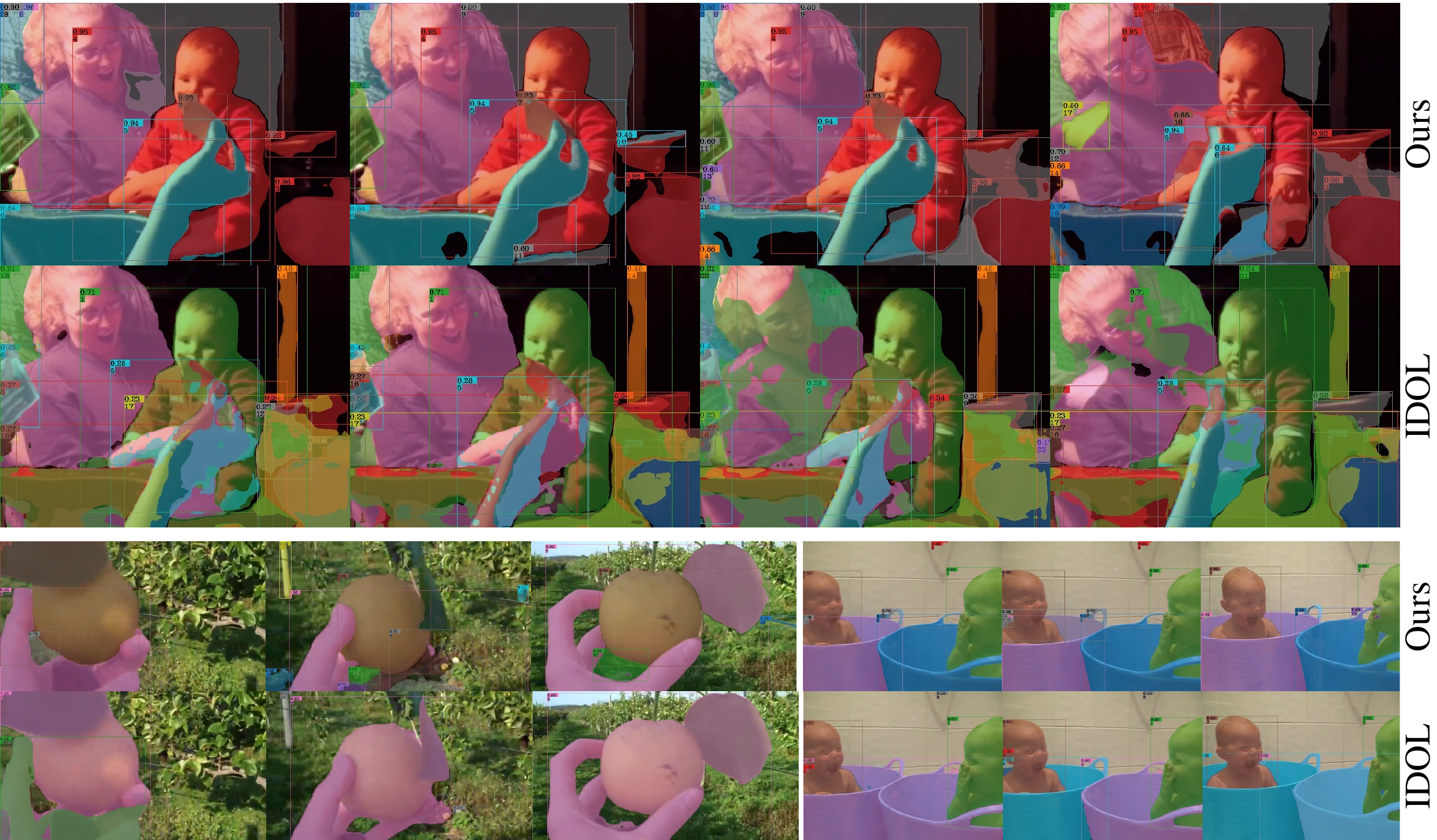}

   \caption{\textbf{Comparison with IDOL on UVO dataset.} Both models use ResNet-50 backbone. ROVIS is consistently able to maintain a better segmentation and distinguish objects near to each other. }
   \label{compare:idol_on_uvo}
\end{figure*}

\begin{figure*}[t]
  \centering
   \includegraphics[width=.95\linewidth]{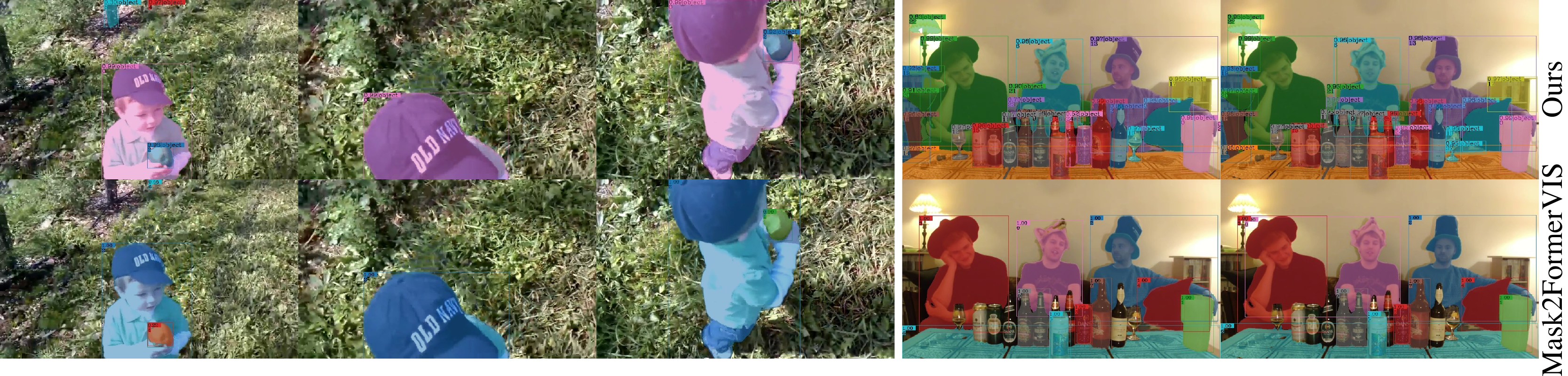}

   \caption{\textbf{Comparison with Mask2Former VIS on UVO dataset.} Both models use Swin-L backbone. \textbf{Left:} Mask2Former VIS makes a tracking error when an object disappears and re-appears (left). \textbf{Right:} In a crowded scene, Mask2Former VIS has high tracking accuracy but fails to find all objects. }
   \label{compare:mask2vis_on_uvo}
\end{figure*}

\noindent{\bf UVO results.}
In Table~\ref{uvo-main}, we compare our model against state-of-the-art methods on the UVO \cite{wang2021unidentified} dataset. UVO has an average video length of 90 frames with many longer videos. Many scenes in UVO are extremely cluttered, thus making it a suitable benchmark for measuring a tracker's linking capability when the number of objects is large. Table~\ref{uvo-main}(a) compares a few methods with a ResNet50 backbone. The baseline result released in the UVO paper used the MaskTrack R-CNN framework \cite{yang2019video} to train an open-world ResNet50 tracker. To form a stronger baseline, we trained the offline Mask2Former VIS model \cite{cheng2021mask2formervis} on UVO using 480px resolution and the same training settings as the original work. This model (evaluated on CPU) gets 19.0 AP, which is slightly higher than our Mask2Former+IoU baseline. We also trained and evaluated IDOL on UVO, since no UVO results were reported in~\cite{IDOL}, and it underperformed both ROVIS and the Mask2Former baselines at 16.8 AP.
Finally, as shown in Table~\ref{uvo-main}(b), the Swin-L version of our model gets 32.7 AP.
Figures \ref{compare:idol_on_uvo} and \ref{compare:mask2vis_on_uvo} show example sequences comparing the output of our method to IDOL and Mask2Former VIS. 
The performance of Mask2Former VIS and IDOL is negatively affected by the cluttered open-world scenario and shape deformations.


\section{Conclusion}

We demonstrated an end-to-end online transformer approach to VIS that carries track information from frame to frame using only lightweight \textit{track query} embeddings, without a need for a separate linking embedding or optimization step. When combined with the strong Mask2Former detector, our ROVIS method matches or outperforms both recent offline and online transformer-based methods on benchmarks with longer videos. Specifically, on the occlusion-rich OVIS dataset, our method outperforms Mask2Former VIS~\cite{cheng2021mask2formervis} and MinVIS~\cite{huang2022minvis} and matches IDOL~\cite{IDOL}, the latter being specifically developed with OVIS in mind. On the challenging class-agnostic UVO dataset, ROVIS decisively outperforms Mask2Former VIS and IDOL. This convincingly demonstrates that the track query idea is robust and versatile enough to cope with occlusions, cluttered scenes, and shape and appearance deformations. Given its efficiency for both training and testing, it makes an attractive starting point for the development of more advanced and high-performing online transformer-based methods.

 


{\small
\bibliographystyle{ieee_fullname}
\bibliography{egbib}
}

\end{document}